\documentclass[10pt,journal,compsoc]{IEEEtran}
\usepackage{url}
\usepackage{amsmath,graphicx}
\usepackage{float}
\usepackage{booktabs}
\usepackage{multirow}
\usepackage{amssymb}
\usepackage{amsmath}
\usepackage[square,sort,sort&compress,numbers]{natbib}
\usepackage[marginal]{footmisc}
\usepackage{lipsum}
\usepackage{algorithm}
\usepackage{algorithmicx}
\usepackage{algpseudocode}
\usepackage[table,xcdraw]{xcolor}
\usepackage[T1]{fontenc}
\newcommand{\norm}[1]{\left\lVert#1\right\rVert}
\title{Manifold-preserved GANs}
\author{Haozhe Liu$^\dagger$, Hanbang Liang$^\dagger$, Xianxu Hou, Haoqian Wu, Feng Liu, Linlin Shen$^*$
\thanks{$^\dagger$Equal Contribution:Haozhe Liu and Hanbang Liang}
\thanks{$^*$Corresponding Author: Linlin Shen: llshen@szu.edu.cn}
\thanks{Haozhe Liu, Hanbang Liang, Xianxu Hou, Haoqian Wu, Feng Liu and Linlin Shen are with the College of Computer Science and Software Engineering, Shenzhen University, Shenzhen 518060, China;
SZU Branch, Shenzhen Institute of Artificial Intelligence and Robotics for Society, China;
Guangdong Key Laboratory of Intelligent Information Processing, Shenzhen University, Shenzhen 518060, China.}
}
\begin{document}
\IEEEtitleabstractindextext{%
\begin{abstract}
Generative Adversarial Networks (GANs) have been widely adopted in various fields. However, existing GANs generally are not able to preserve the manifold of data space, mainly due to the simple representation of discriminator for the real/generated data. To address such open challenges, this paper proposes Manifold-preserved GANs (MaF-GANs), which generalize Wasserstein GANs into high-dimensional form. Specifically, to improve the representation of data, the discriminator in MaF-GANs is designed to map data into a high-dimensional manifold. Furthermore, to stabilize the training of MaF-GANs, an operation with precise and universal solution for any $K$-Lipschitz continuity, called Topological Consistency is proposed. The effectiveness of the proposed method is justified by both theoretical analysis and empirical results. When adopting DCGAN as the backbone on CelebA (256$\times$256), the proposed method achieved 12.43 FID, which outperforms the state-of-the-art model like Realness GAN (23.51 FID) by a large margin. \textit{Code will be made publicly available.}
\end{abstract}
}
\maketitle
\section{Introduction}
Generative Adversarial Networks (GANs) have been widely applied in various tasks \cite{creswell2018generative,8411144,8968618,9149832,9496081,liu2021group}. As a density estimation tool, a generator and a discriminator are used to estimate the underlying distribution of the given data. The generator models the data distribution by mapping prior distribution to a generated data distribution, while the discriminator learns to judge whether the input data is from the generated or real data distribution. Taking a min-max game as the objective, the discriminator leans to distinguish generated samples from real data distribution by maximizing distance between two distributions, and the target of generator is to spoof discriminator by minimizing the distance .

As GANs are trained in an unsupervised mode where the optimization of the generator only depends on the state of the discriminator, the properties of discriminator are of vital importance for the generative performance of the GANs.
Based on such observation, an ideal solution to facilitate GANs is to employ a discriminator with strong representation and continuous weights. For the optimization of the generator, progressive gradients can be contributed by discriminative representation that lands on the proper direction, and the continuity of the discriminator which provides a stable training process. The solid evidences of this point are Wasserstein GAN (WGAN) \cite{arjovsky2017wasserstein} and WGAN-GP \cite{gulrajani2017improved}. WGAN
introduces wasserstein distance into the framework of the GANs, and thus improves the representation of the discriminator by transferring the classification-like objective to the regression-like objective. In terms of continuity, an operation, called weight clipping, is adopted in WGAN to ensure the 1-Lipschitz continuity.
However, extensive empirical studies \cite{gulrajani2017improved,wu2018wasserstein,stanczuk2021wasserstein} show that, due to the sensitivity to the clipping value, the weight clipping is not a reasonable way to provide stable and progressive gradients. To tackle this kind of problems, WGAN-GP is proposed to adopt gradient penalty as the regularization for stable training.
As a well-known and useful solution, WGAN-GP is widely used in different GANs, and gradually becomes a common loss for GANs  \cite{karras2017progressive,karras2019style,choi2018stargan}.

Despite of the tremendous success of WGAN-GP, there remains unsolved problems and unexplained phenomenons for the properties of the discriminator. When it comes to the discriminative representation, the discriminator in the family of WGANs generally maps data to a single scalar, which could be viewed as an abstraction of the generated and real data distribution. However, the samples are high dimensional data with multiple attributes, which can not be represented comprehensively through low dimensional space.
As a result, generator could easily get stacked in local optima, resulting in issues such as mode-collapse and gradient exploding. Hence, \textbf{generalizing the representation to an appropriate embedding space is crucially important for the family of WGANs}.
On the other side, in the terms of continuity, the weights of WGAN is constrained by a pre-defined bound to lie within a compact space, while WGAN-GP penalizes the norm of gradient of the discriminator with respect to its input. Although the general idea of both methods is identical, i.e. limiting the parameter space of discriminator, quite different generative results are achieved by WGAN and WGAN-GP.  It seems that, \textbf{limited parameter space is not an ideal constraint for the continuity, and the fundamental reason for the advance of WGAN-GP is still unclear.}

To address the mentioned open problems, this paper firstly redefines the family of WGANs as Manifold-preserved GANs (MaF-GANs), and generalizes the traditional wasserstein loss to high dimensional forms. Based on the high dimensional representation, the strategy to ensure 1-Lipschitz continuity is revised carefully. Unlike other approximation, the proposed method gives a \textbf{precise bound}, denoted as Topological Consistency, for the Lipschitz constraint with theoretical proof. By adopting Topological Consistency to analyze the continuity, this paper demonstrates that the continuity of weight-clipping fluctuates in a large region, while that of WGAN-GP is stable. Such result demonstrates the radical differences among existing continuity constraints and further proves that limited parameter space is insufficient for the continuity.

To sum up, our contribution can be concluded as follows:
\begin{itemize}
    \item This paper proposes Manifold-preserved GANs, where the discriminator can better represent any given data, and provide more precise optimization direction by mapping data into high dimensional embedding space.
    \item Without any approximations, the precise solution of Lipschitz continuity, namely Topological Consistency, is derived and proved mathematically.
    \item The rational for the advance of WGAN-GP over WGAN is clearly given in this paper, which further demonstrates the superiority of Topological Consistency.
    \item Experimental results show that, based on a rather simple DCGAN architecture, the proposed Maf-GANs can achieve 12.43 Fr\'{e}chet Inception Distance (FID), which is much lower than that of realness GAN (23.51).
\end{itemize}

\section{Background}
\begin{table}[!htbp]
\centering
\caption{The Summary of the Adopted Notations and the Corresponding Explanation}
\setlength\tabcolsep{5pt}
 \resizebox{.5\textwidth}{!}{
\begin{tabular}{cc}
\hline
Notation & Explanation \\\hline
$z$      & the input vector of the generator           \\
$x_g$             & the generated data            \\
$x_r$             & the real data            \\
$\mathbf{V}_x$      &  the discriminative embedding  of $x$ \\ \hline \hline
$D(\cdot)$             & the discriminator in GANs           \\
$G(\cdot)$             & the generator in GANs            \\
$f(\cdot)$             & the mapping function to process $\mathbf{V}_x$ \\ \hline\hline
$p_r$             & the real data distribution            \\
$p_{z}$               & the distribution of $z$ \\
$p_{g}$               & the distribution of the generated samples \\
$\mathbb{U}(0,1)$      & the uniform distribution from 0. to 1. \\
$\mathbb{R}^n$         & N-dimensional real number \\\hline \hline
$\mathbb{L}(\cdot,\cdot)$           & the distance metrics applied in min-max game \\
$\mathcal{L}$                  & the learning objectives for GANs \\
$\mathbb{D}_{TC}$                        & Topological Consistency proposed in this paper \\
\hline
\end{tabular}
 }
\end{table}
\subsection{The Learning Objectives for GANs}
In GANs, a generator $G(\cdot)$ and a discriminator  $D(\cdot)$ learn in an adversarial manner. $G(\cdot)$ maps a vector $z$ from the prior $p_z$ to a generated data $x_g$ obeying to distribution $p_{g}$. On the other hand, $D(\cdot)$ distinguishes $x_g$ from real data $x_r$. As a competitive game, the distance between $x_g$ and $x_r$ is maximized and minimized respectively to train $D(\cdot)$ and $G(\cdot)$. Formally, such game can be defined as
\begin{equation}
\label{general_gan}
    \min_{G} \max_{D} \underset{\substack{z \sim p_z \\x \sim p_r}}{\mathbb{L}}\!(f(D(x)), f(D(G(z))))
\end{equation}
where $p_r$ is the data distribution, $f(\cdot)$ refers to a mapping function $f : \mathbb{R}^n \rightarrow \mathbb{R}^m$ and $\mathbb{L}$ is a metric. Note that, the output of $D(x)$ is regarded as a discriminative embedding code (vector) $\mathbf{V}_x$, rather than a single scalar.
Standard GAN (Std-GAN) \cite{goodfellow2014generative} is drawn as a subset of the mentioned objective when
\begin{equation}
\label{vanilagan}
\begin{aligned}
   \underset{\substack{z \sim p_z \\x \sim p_r}}{\mathbb{L}_{std}}(f(D(x))&, f(D(G(z)))) = \underset{x \sim p_r}{\mathbb{E}}[\log f(D(x))] \\
   &+ \underset{z \sim p_{z}}{\mathbb{E}}[\log(1-f(D(G(z))))]
\end{aligned}
\end{equation}
where $\mathbb{L}(\cdot,\cdot)$ is defined as cross entropy and $f$ is implemented as a fully-connected layer with sigmoid.

\subsection{Lipschitz Continuity based Wasserstein GANs}
When training GANs by Eq.(\ref{vanilagan}), the target of $D(\cdot)$ is transformed to minimize the Jensen-Shannon (JS) divergence between $p_r$ and $p_g$. As JS divergence discretely evaluates the differences, the progressive gradient of $G(\cdot)$ is hard to be derived from $D(\cdot)$. To circumvent this difficulty, many studies \cite{mao2017least,arjovsky2017wasserstein,gulrajani2017improved,zhao2016energy} use the different metrics to quantify the distance between $p_r$ and $p_g$. Based on the reported results, one of the most practical solutions is  WGAN, which guides $G(\cdot)$ to spoof $D(\cdot)$ in a continuous way.

In WGAN. wasserstein (also called Earth-Mover) distance is used as metric, which is informally defined as the minimum cost of transporting mass in order to transform one distribution to another. Under mild assumptions, wasserstein loss is continuous and differentiable everywhere. By applying the Kantorovich-Rubinstein duality \cite{villani2009optimal}, WGAN can be expressed as:
\begin{equation}
 \label{wgan}
    \min_{G} \max_{D\in\mathcal{D}} \underset{x \sim p_r}{\mathbb{E}}[f^1(D(x))]-\underset{z \sim p_{z}}{\mathbb{E}}[f^1(D(G(z)))]
\end{equation}
where $\mathcal{D}$ is the set of 1-Lipschitz functions and $f^1: \mathbb{R}^n \rightarrow \mathbb{R}^1 $, refers to a fully-connected layer without any activators. Unlike other GANs trained by classification-like objective, WGAN regards the generative task as a regression-like problem, making optimization of the generator easier. However, such transformation introduces a strong constraint for $D(\cdot)$, i.e. Lipschitz continuity, several studies thus propose some approximations to meet the requirement. Arjovsky et al. \cite{arjovsky2017wasserstein} clip the weight of $D(\cdot)$ within a compact space [-c,c]. Gulrajani et al. \cite{gulrajani2017improved} penalize the norm of gradient of $D(\cdot)$ with respect to the given samples.

\begin{figure*}[!htbp]
    \centering
    \includegraphics[width=0.98\textwidth]{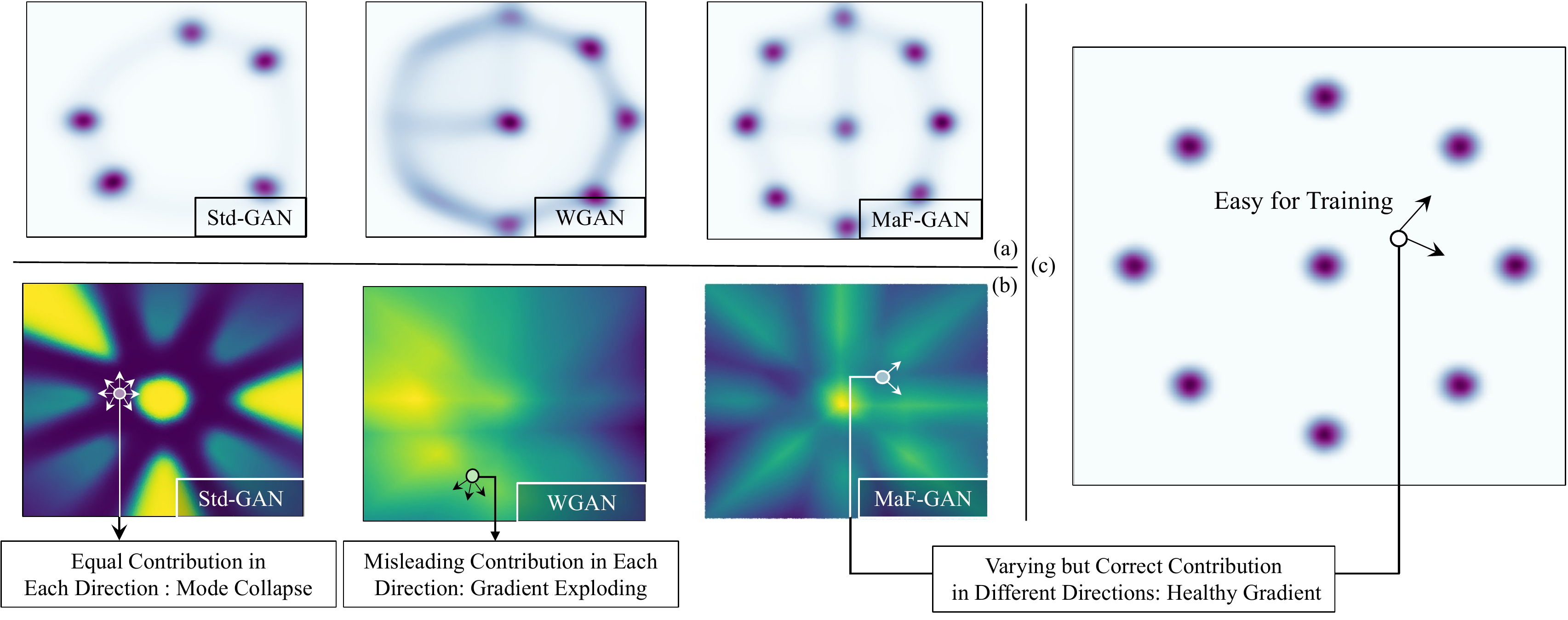}
    \caption{A toy example, where training set is composed of the 2D points from 9 mixed Gaussian distributions. (a) From left to right: the samples generated by Std-GAN, WGAN and MaF-GAN (Ours). (b) The corresponding confidence maps of the discriminator: the point with higher value indicates that the discriminator classifies the point into real data distribution with stronger confidence. (c) The points sampled from real data distribution. }
    \label{fig:activation_map}
\end{figure*}
\begin{figure*}[!htbp]
    \centering
    \includegraphics[width=0.98\textwidth]{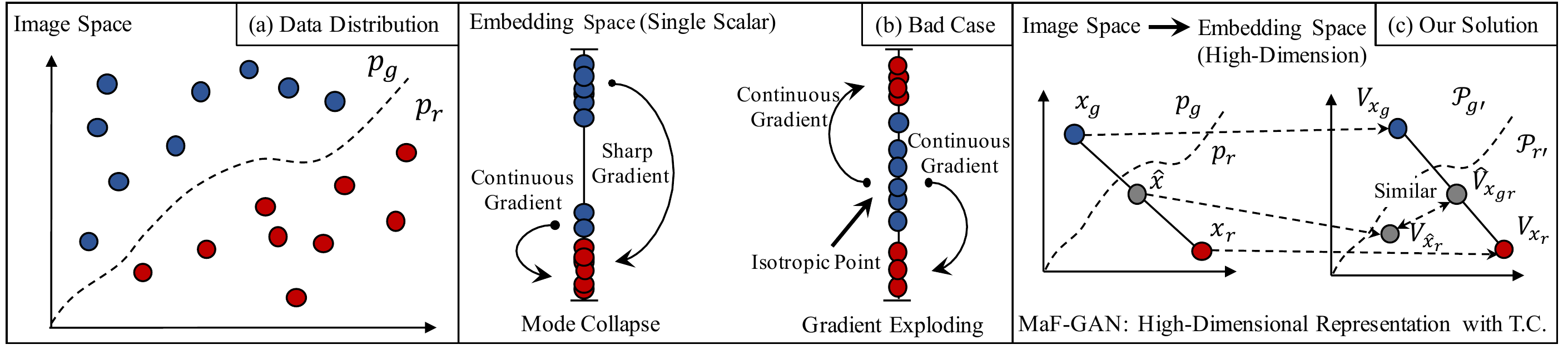}
    \caption{Some learning objectives for GANs. (a) refers to the data distribution, (b) presents the bad cases of the existing learning objectives and (c) shows the proposed solution, which trains GANs with high-dimensional representation. In (a), generated data distribution $p_g$ and real data distribution $p_r$ are supported in the image space. Existing GANs generally maps $p_g$ and $p_r$ into a single scalar (i.e. a line shown in (b)) may lead to the destruction of the topological structure resulting in some unexpected mode such as mode collapse and gradient exploding. To mitigate the problem, the proposed method embeds $p_g$ and $p_r$ into $p_{g'}$ and $p_{r'}$ which stand in a high-dimensional latent space spanned by $\mathbf{V}_x$. With strong representation, Topological Consistency is designed for comprehensive manifold preservation.}
    \label{fig:badcase}
\end{figure*}
\subsection{Mode collapse and gradient exploding}
As summarized from the aforementioned studies, whether regression-like or classification-like objectives, the applied $f(\cdot)$ is similar, i.e. maps $x$ to a single scalar. However, considering the complexity of $p_g$ and $p_r$, the capacity of the scalar in representing $x$ is very limited, which might lead to some undesired behaviors. In order to demonstrate some of the issues caused by poor representation, some phenomenons observed in a toy example are carefully discussed, which further motivate the proposed method.

In this toy example, we train different GAN models, using 50k points generated from 9 mixed Gaussian distributions. In these models, $G(\cdot)$ consists of 4 fully-connected hidden layers and  $D(\cdot)$ is composed of three fully-connected layers. To guarantee the fairness, hyper-parameters are identical for different learning objectives. As shown in Fig. \ref{fig:activation_map}(a),  the samples generated by Std-GAN only rest on partial of the distribution, which is a typical example of mode collapse. To understand such phenomenon, we visualize $f(D(\cdot))$ with respect to each point in the generative space and denote it as confidence map. The higher value confidence map shows, the more confident $D(\cdot)$ judges the selected point as a real sample from $p_{data}$. Since the target of $G(\cdot)$ is to spoof $D(\cdot)$, the gradients of $G(\cdot)$ shall move $G(\cdot)$ from low to high confidence value.
However, the confidence map in Fig. \ref{fig:activation_map}(b), shows that only the points in Gaussian distribution are with high confidence, while other regions present similarly low confidence. This indicates that $G(\cdot)$ derives binary gradients from $D(\cdot)$, i.e., the realness of a given point is only 0. or 1.
In the view of $G(\cdot)$, the points with different distances to the real distribution share the similar confidence, hence, it is an over-heavy cost for $G(\cdot)$ to generate adjacent samples surrounding the local distribution with  high confidence, leading to the trap of local optima, i.e., mode collapse.

As shown in Fig. \ref{fig:activation_map}(a), while some of the samples generated by WGAN follow five of the nine Gaussian distributions, other samples are diversely distributed along the circle and do not form any clusters. Compared with Std-GAN, the confidence map of WGAN is continuous and will not lead to mode collapse. However, as indicated by the example in Fig. \ref{fig:activation_map}(b), the gradients of optimization at many locations are isotropic, i.e. no dominant direction is available, which may cause convergence problem and even lead to gradient exploding.

Fig.\ref{fig:badcase} further shows the differences between the learning objectives of current GANs and our proposed MaF-GAN. The current GANs usually embed the generated samples into a scalar to indicate how real the samples are, which actually can not well preserve the topological structures of $p_g$ and $p_r$. As shown in Fig. \ref{fig:badcase}(b), when the generated samples are discretely distributed, i.e. some are close to $p_g$ and others are far away from $p_r$, mode collapse occurs. When the generated samples are continuously distributed and the gradients are isotropic i.e. no dominant direction for optimization is available, gradient exploding occurs. On both cases, $D(\cdot)$ cannot provide appropriate feedback required by $G(\cdot)$ to improve the quality of generated samples.

As topological structure is the crucial factor to tackle the aforementioned problems, this paper proposes two measurements to preserve the manifold of $p_g$ and $p_r$. Firstly, the proposed method maps $p_g$ and $p_r$ to a high-dimensional latent space rather than a single scalar. Since more dimensions are adopted for representation against information loss, more comprehensive description of $p_g$ and $p_r$ are preserved in the embedding space.
To measure the difference between $p_g$ and $p_r$ in such an embedding space, the expectation, pivot and cosine similarity based metrics are discussed carefully in this paper. Secondly,  a Topological Consistency is proposed to preserve the topological structure between $p_g$ and $p_r$. Based on the consistency, $D(\cdot)$ is explicitly required to preserve the operational identity in both image space and embedding space. More theoretically, this paper gives a formal proof for the effectiveness of Topological Consistency: such consistency is the precise bound for Lipschitz constraint. Meanwhile, by taking Topological Consistency as an analysis tool, the relationship between the manifold and the generated result can be observed explicitly, which further explains the effect of Lipschitz continuity on GANs.

\section{Manifold-preserved Discriminator}
Unlike existing methods, the proposed method, denoted as MaF-GAN, is quite different from the conventional GANs. As shown in Fig. \ref{fig:badcase}(b), previous GANs generally embed the generated and real distribution into a single scalar, which might easily lead to unstable training and then result in serious mode collapse and gradient exploding. To address such problems, MaF-GAN (see Fig.\ref{fig:badcase}(c)) maps the given samples into high-dimensional embedding space and design a strong constraint, denoted as Topological Consistency, to preserve the topological structure. In the following subsections, we will give the detailed presentation of the proposed method.
\subsection{The Zoo of  MaF-GANs}
The discriminator of GANs usually processes the input in two stages: Firstly. $D(\cdot)$ embeds input $x$ into a vector $\mathbf{V}_x$, and secondly, applies a mapping function $f(\cdot)$ to process $\mathbf{V}_x$. Existing GANs generally employ a series of fully-connected layers as $f(\cdot)$ to compress $\mathbf{V}_x$ into a scalar, however such  over-compression may lead to poor representation for $x$,
resulting in some unexpected modes, such as mode-collapse and gradient exploding.

To tackle the problem, MaF-GAN modifies $f^m(\cdot):  \mathbf{R}^n \rightarrow \mathbf{R}^m$ as a space mapper, which directly transforms the $n$-dimensional space (supporting for $p_g$ and $p_r$) into the embedding space spanned by $\mathbf{V}_x$ with $m$ dimensions.
Among $\mathbf{V}_x$, each element refers to an independent metric to evaluate the distance between $p_g$ and $p_r$. Through  $\mathbf{V}_x$, more comprehensive differences between $p_g$ and $p_r$ can be observed. Since  $\mathbf{V}_x$ is quite different from a single scalar, we consider three potential solutions, including MaF-CGANs, MaF-DGANs and MaF-$\mathbb{E}$GANs, to analyze $\mathbf{V}_x$.
\subsubsection{MaF-CGANs: Cosine Similarity based MaF-GANs}
In cosine similarity based MaF-GANs (namely MaF-CGANs), we introduce an auxiliary parameter $\mathbf{W} \in \mathbf{R}^m$ as a pivot to estimate the representative embedding of $x_r$. By minimizing and maximizing the similarity between $\mathbf{V}_x$ and $\mathbf{W}$ respectively, $G(\cdot)$ and $D(\cdot)$ can be trained in a high-dimensional latent space. Specifically, cosine similarity $\mathbb{L}_{C.}(\mathbf{V}_x,\mathbf{W})$ is defined to evaluate the difference between $\mathbf{V}_x$ and $\mathbf{W}$,
\begin{align}
        \mathbb{L}_{C.}(\mathbf{V}_x,\mathbf{W})=\frac{\mathbf{V}_x \cdot \mathbf{W}}{\|\mathbf{V}_x \|\|\mathbf{{W}}\|}=\frac{\sum_{i}^{N} {\mathbf{V}_x}_{i} {W}_{i}}{\sqrt{\sum_{i}^{m} {\mathbf{V}_x}_{i}^{2}} \sqrt{\sum_{i}^{m} {W}_{i}^{2}}}
\end{align}
where $i$ is the $i$th element for $\mathbf{V}_x$ or $\mathbf{W}$. Given the real data $x_r$ as input, the representative embedding of $x_r$ can be obtained by maximizing $\mathbb{L}_{C.}(\mathbf{V}_{x_r},\mathbf{W})$. The goal of $G(\cdot)$ is to generate samples surrounding $\mathbf{W}$, while $D(\cdot)$ aims to discriminate $x_g$ from $\mathbf{W}$. Considering $\mathbf{W}$ as a trainable parameter, the learning objective of MaF-CGANs can be concluded as,
\begin{align}
     \min_{G} \max_{D, \mathbf{W}} \!\underset{x \sim p_r}{\mathbb{E}}\!\![\mathbb{L}_{C.}(f(D(x)),\mathbf{W})]\!-\!\!\!\!\!\underset{z \sim p_{z}}{\mathbb{E}}\!\![\mathbb{L}_{C.}(f(D(G(z))),\mathbf{W})]
\end{align}

\subsubsection{MaF-DGANs: Prior Distribution based MaF-GANs}
Defining a high-dimensional distribution $\mathcal{Q}$ as a prior of $p_r$, the probability of $\mathbf{V}_x$ obeying to $p_r$ can be measured by the probability density $\mathcal{Q}(\mathbf{V}_x)$. $D(\cdot)$ tries to discriminate $p_r$ and $p_g$ using $\mathcal{Q}(\mathbf{V}_x)$, while the target of $G(\cdot)$ is contrary to that of $D(\cdot)$. In such manner, the metric $\mathbb{L}_{D.}(\mathbf{V}_x)$ is expressed as,
\begin{equation}
\label{LD}
\begin{aligned}
    \mathbb{L}_{D.}(\mathbf{V}_x) &= \mathcal{Q}^*(\mathbf{V}_x) \\
    s.t.\quad  \mathcal{Q}^* &= \mathop{\arg\max}_{\mathcal{Q}} \underset{x_r \sim  p_r}{\mathbb{E}}[\log(\mathcal{Q}(V_{x_r}))]
\end{aligned}
\end{equation}
where $\mathcal{Q}^*$ is the optimal prior distribution for $p_r$, which can be optimized by maximizing likelihood. Based on $\mathcal{Q}^*$, $G(\cdot)$ and $D(\cdot)$ are alternatively trained by minimizing and maximizing $\mathbb{L}_{D.}$ respectively. The learning objective can be concluded as,
\begin{equation}
    \min_{G} \max_{D} \underset{x \sim p_r}{\mathbb{E}}[\mathbb{L}_{D.}(f(D(x)))]-\underset{z \sim p_{z}}{\mathbb{E}}[\mathbb{L}_{D.}(f(D(G(z))))]
\end{equation}
Various options can be selected as a prior distribution. In our case, standard Gaussian distribution is adopted as $\mathcal{Q}$.
\subsubsection{MaF-$\mathbb{E}$GANs: Expectation based MaF-GANs}
\label{sec:EGAN}
In MaF-$\mathbb{E}$GANs, $\mathbf{V}_x$ is regarded as a joint distribution, hence the critic value of $p_r$ and $p_g$ can be directly calculated by estimating the expectation of $\mathbf{V}_x$.
In particular, the metrics $\mathbb{L}$ of MaF-$\mathbb{E}$GANs combines a norm term and an entropy term to enforce each element in $\mathbf{V}_x$ to be an independent critic value:
\begin{equation}
\label{maf-egans}
\mathbb{L}_{\mathbb{E}}(\mathbf{V}_x) = \frac{\norm{{\mathbf{V}_x}}_{1}}{m} - \mathbb{E} \log \mathbf{V}_x
\end{equation}
where $\norm{\cdot}_1$ represent L-1 norm. The key idea behind MaF-$\mathbb{E}$GANs is to simulate multiple $f^1$ discriminators for each element of $\mathbf{V}_x$. However, simply applying element-wise mean operation over $\mathbf{V}_x$ would lead to a trivial solution, where each element of $\mathbf{V}_x$ has the same value, i.e. $f^m(\cdot)$ degenerates into  $f^1(\cdot)$. To address this, an entropy term is added to force each element of $\mathbf{V}_x$ to learn more comprehensive representation. Similarly to MaF-CGANs and MaF-DGANs, the learning objective can be presented as:
\begin{align}
    \min_{G} \max_{D} \underset{x \sim p_r}{\mathbb{E}}[\mathbb{L}_{\mathbb{E}}(f(D(x)))]-\underset{z \sim p_z}{\mathbb{E}}[\mathbb{L}_{\mathbb{E}}(f(D(G(z))))]
\end{align}
\subsection{Training MaF-GANs with Topological Consistency}
For the first time, MaF-GANs generalize Wasserstein distance into high dimensional embedding space to train GANs. As the dimension scales up, the representation capability of data $x$ expands enormously over WGANs, which also brings new challenges such as trivial solution and unstable training.
In MaF-GANs, trivial solution maps  $\mathcal{C}$ data points into the same $\mathbf{V}_x$, i.e. $D(\cdot): \mathcal{C} \rightarrow 1 $. Such manner severely destructs the topological structure among images and suppresses the diversity of representation, which might further leads to hard training.

To mitigate this problem, this paper proposes a strong constraint  $\mathbb{D}_{TC}$, denoted as Topological Consistency, for $\mathbf{V}_x$. The premise of the consistency is that when mapping $x_r$ and $x_g$ into a compact space, topological structure between samples should be preserved. Specifically, two samples $x_r$ and $x_g$ are mixed up through an operator $\Delta(\cdot, \cdot)$
\begin{align}
     \Delta(x_g, x_r)_{\epsilon \sim \mathbb{U}(0,1)} =\epsilon  x_r + (1-\epsilon)  x_g
\end{align}
where $\mathbb{U}(0,1)$ refers to the uniform distribution between [0, 1) and $\epsilon$ is a scalar sampled from $\mathbb{U}(0,1)$. Based on $\Delta(\cdot, \cdot)$, Topological Consistency for $p_g$ and $p_r$ can be formally defined as
\begin{align}
     \mathbb{D}_{TC}&=\underset{\substack{z \sim p_z \\x_r \sim p_r }}{\mathbb{E}} [d(\mathbf{V}_{\hat{x}} , \mathbf{\hat{V}}_{x_{gr}} ) + \delta ] \\
     \mathbf{V}_{\hat{x}} &= D(\hat{x}) = D(\Delta(x_g, x_r)) \\
     \mathbf{\hat{V}}_{x_{gr}} &= \Delta(\mathbf{V}_{x_r},\mathbf{V}_{x_g} )
\end{align}
where $d(\cdot,\cdot)$ is a metric to evaluate the difference between $\mathbf{V}_{\hat{x}}$ and $\Delta(\mathbf{V}_{x_r},\mathbf{V}_{x_g} )$ and $\delta$ is a random perturbation to ensure the progressive gradient and stable training. In this paper, mean square error is adopted as $d$ and $\delta$ is sampled from a Gaussian distribution with 0 mean and 0.05 standard deviation. Following $\mathbb{D}_{TC}$, $D(\cdot)$ is required to preserve identical Topological Consistency through data space and embedding space. The final learning objective is defined as
\begin{equation}
        \min_{G} \max_{D \in \mathbf{D}} \underset{\substack{z \sim p_z \\x \sim p_r}}{\mathbb{L}}\!(f(D(x)), f(D(G(z))))
\end{equation}
where $\mathbf{D}$ is the set of $D(\cdot)$ satisfying $\mathbb{D}_{TC}$.
Since $\mathbb{D}_{TC}$ is a constraint for the embedding space spanned by $\mathbf{V}_x$, $G(\cdot)$ is not trained through $\mathbb{D}_{TC}$. For clarity, the proposed method is summarized as Algo.1
\begin{algorithm}[!htb]
  \label{algo:2}
	\caption{Manifold GAN with Topological Consistency}
	\begin{algorithmic}
    \Require\\
     Generator $G_{\theta_0}(\cdot)$; Discriminator $D_{\gamma_0}(\cdot)$; Manifold Operation $f(\cdot)$; The number of critic iterations per generator iteration $n_{\text{critic}}$
    \Ensure \State Trained Parameters $\theta$;
  \end{algorithmic}
  \begin{algorithmic}[1]
    \While{ $\theta$ has not converged}
         \For{$t=1$ to $n_{\text{critic}}$ }
            \State Sample real data $x_r \sim p_r$, latent variable $z \sim P_z$;
            \State $ x_g \leftarrow G_{\theta}(z)$;
            \State $\hat{x} \leftarrow \Delta(x_r,x_g) $;
            \State $\mathcal{L} \leftarrow \mathbb{L}( f(D_{\gamma}(\widetilde{x}))) -  \mathbb{L}(f(D_{\gamma}(x)))$;
            \State $ \mathbb{D}_{TC} = d(D_{\gamma}(\hat{x}), \Delta(D_{\gamma}(x) , D_{\gamma}(\widetilde{x})))$;
            \State $\gamma_t \leftarrow$ Adam($\frac{\partial (\mathcal{L} + \mathbb{D}_{TC})}{ \partial {\gamma_{t-1}}}$);
         \EndFor
         \State Sample latent variable $z \sim P_z$;
         \State $\mathcal{L} \leftarrow \mathbb{L}(f(D(G_\theta(z))))$;
         \State $\theta \leftarrow$ Adam($-\frac{\partial \mathcal{L}}{ \partial \theta}$);
 \EndWhile
\State Return $\theta$;
	\end{algorithmic}
\end{algorithm}

\section{Theoretical Analysis: The Equivalence between Lipschitz Continuity and Topological Consistency}
\label{sec:ta}
To further investigate the effectiveness of the proposed method, we theoretically analyze Topological Consistency and show the equivalence between Lipschitz continuity and the proposed consistency.

\textbf{Lipschitz Continuity in GANs}: Given two metric spaces $(\mathcal{N},d_n)$ and $(\mathcal{M},d_m)$, where $d_n$ denotes the metric on the set $\mathcal{N}$, and $d_m$ is the metric on set $\mathcal{M}$. A function $f(D(\cdot))$: $ \mathcal{N} \rightarrow \mathcal{M}$ is called Lipschitz continuity if there exits a real constant $K$ such that, for all $x_r \sim p_r$ and $x_g \sim p_g$ in $\mathcal{N}$,

\begin{equation}
\label{eq:lip}
    d_n(D(x_r),D(x_g)) \leq K d_m(x_r,x_g), s.t., x_r,x_g \in \mathcal{N}
\end{equation}

where $d_n(\cdot,\cdot) \geq 0 $.

\textbf{Proposition 1.} Based on Kantorovich-Rubinstein duality \cite{villani2009optimal}, K-Lipschitz function set $\mathcal{D}$ is the optimal solution of $\max_{D \in \mathcal{D}} K \mathbb{L}(\mathbf{V}_{x_r}, \mathbf{V}_{x_g}) $.  Since $K$ is an absolute term for the objective of $D(\cdot)$, the optimal result of $D(\cdot)$ is identical among any given $K$.

\textbf{Corollary 1.} As $\mathbb{L}(\mathbf{V}_{x_r}, \mathbf{V}_{x_g})$ is different among different tasks and data distributions, $\mathcal{D}$ should be the set of universal solutions for any K-Lipschitz constraints to stabilize the training of $D(\cdot)$. Existing methods, such as gradient penalty and weight clipping can not meet such requirement.

\textbf{Theorem 1.} $K$-Lipschitz Continuity of $D(\cdot)$ can be ensured by the optimal result of Topological Consistency.

\textbf{Hypothesis} $d_n(\cdot,\cdot)$ and $d_m(\cdot,\cdot)$ are the linear metrics,
\begin{equation}
\begin{aligned}
    \alpha d_m (x_1,x_2) + \beta d_m (x_3,x_4) &= \\
    d_m(\alpha x_1 &+ \beta x_3, \alpha x_2 + \beta x_4)
\end{aligned}
\end{equation}
\begin{equation}
\begin{aligned}
   \alpha d_n (D(x_1),D(x_2)) + \beta d_n (D(x_3),D(x_4)) &= \\
   d_n(\alpha D(x_1) + \beta D(x_3), \alpha D(x_2) + \beta &D(x_4))
\end{aligned}
\end{equation}
where the set $\{x_i|i=1,2,3,4 \}$ is randomly sampled from $\mathcal{N}$.

\textbf{Proof.} Given $x \sim p_r$ and $G(z)$ with $z \sim P_z$ as input, three linear combinations, $C_1$, $C_2$, $C_3 \in \mathcal{N}$, are defined as,
\begin{gather}
    C_1 = \mathbb{E}_{x \sim p_r}[\epsilon x] + \mathbb{E}_{z \sim p_{z}}[(1-\epsilon) G(z)] \\
    C_2 = \mathbb{E}_{x \sim p_r}[\epsilon x]\\
    C_3 = \mathbb{E}_{z
    \sim p_{z}}[(1-\epsilon) G(z)]
\end{gather}
where $\epsilon$ is a random parameter to mix up two samples, which is introduced in Topological Consistency. Since $C_1$, $C_2$, $C_3$ $\in$ $\mathcal{N}$, the linear combinations,  $C_1$ and $C_2 + C_3$ should also keep within the Lipschitz bound. Through Eq.(\ref{eq:lip}), Lipschitz continuity can be represented as,
\begin{equation}
\begin{aligned}
&d_n(D(C_1),D(C_2)+D(C_3)) = \\
&d_n(D(C_1),D(C_2)) + d_n(D(C_1), D(C_3)) - d_n (D(C_1), O)\\
&\leq K d_m(C_1,C_2) + K d_m(C_1,C_3) - K d_m(C_1,0) \\
&= K d_m(C_1,C_2+C_3)\\
&= K \cdot 0 = 0
\end{aligned}
\end{equation}
where $O$ means the origin of $\mathcal{N}$ or $\mathcal{M}$.
Since $ d_n(\cdot,\cdot) \geq 0$, the unique solution is
\begin{equation}
d_n(D(C_1),D(C_2)+D(C_3)) = 0
\end{equation}
which is identical with the optimal result of Topological Consistency $\mathbb{D}_{TC}$ when $d_n(\cdot,\cdot)$ is adopted as $d(\cdot,\cdot)$ in MaF-GANs. Such equivalence theoretically proves that without any approximations, Topological Consistency is the supremum of Lipschitz continuity for any given K. Note that, since $\mathbb{D}_{TC}$ is the universal solution for all $K \in \mathbf{R}^1$, $\mathbb{D}_{TC}$ is more robust to different hyper-parameters and various data distribution. Meanwhile, $\mathbb{D}_{TC}$ can reflect the continuity of $D(\cdot)$,  $\mathbb{D}_{TC}$ is also a precise metric to evaluate the degree of Lipschitz constraint.

\textbf{Special Case} To simplify the proof, $d_n$ and $d_m$ is assumed as two linear metrics. However, other famous non-linear metrics can also meet the requirement. In this paper, we take mean square error as a special case to further prove the flexibility of the proposed method. Specifically, $d_n(\cdot,\cdot)$ is Mean Square Error (Non-Linear Metric) and $d_m(\cdot,\cdot)$ is the linear metric,
\begin{equation}
\begin{aligned}
 \alpha d_m (x_1,x_2) + \beta d_m (x_3,x_4) &= \\
 d_m(\alpha x_1+ \beta x_3, & \alpha x_2 + \beta x_4)
\end{aligned}
\end{equation}
\begin{equation}
\begin{aligned}
    d_n (D(x_1),D(x_2)) &= \\
    D^2(x_1) - 2 &D(x_1)D(x_2) + D^2(x_2)
\end{aligned}
\end{equation}
The Lipschitz continuity can be explained as,
\begin{equation}
\begin{aligned}
&d_n(D(C_1),D(C_2)+D(C_3))\\
=& d_n(D(C_1),D(C_2)) + d_n(D(C_1),D(C_3)) - \\
& d_n(D(C_2),D(C_3))- d_n(D(C_1),0) + \\
& d_n(D(C_2),0) + d_n(D(C_3),0)  \\
\leq& K d_m(C_1,C_2) + K d_m(C_1,C_3) - K d_m(C_2,C_3) - \\
&K d_m(C_1,0) + K d_m(C_2,0) + K d_m(C_3,0) \\
= &K d_m(C_1-C_2,C_3)\\
= &K \cdot 0 = 0
\end{aligned}
\end{equation}
where the result is identical with the case that $d_n(\cdot,\cdot)$ is linear metric.

\section{Experimental Results and Analysis}
To evaluate the performance of the proposed method, extensive experiments are carried on different generative datasets with different resolutions, including a synthetic dataset (2*1), CIFAR10 (32*32) \cite{krizhevsky2009learning} and CelebA(256*256) \cite{liu2015deep}. In this section, we firstly introduce the data sets and implementation details. Then, the effectiveness of the proposed method is proved by discussing the contribution of each components. Subsequently. by adopting Topological Consistency as an analysis tool to evaluate the continuity of $D(\cdot)$, we carefully investigate the differences among the Wasserstein distance based GANs with different implementations for K-Lipschitz constraint. Finally, the improvement of the proposed methods are justified by comparing with the state-of-the-art learning objectives.
\subsection{Datasets and Implementation Details}
We evaluate the proposed method on three datasets, including a synthetic dataset (2*1), CIFAR10 (32*32) \cite{krizhevsky2009learning} and CelebA(256*256) \cite{liu2015deep}.

\begin{figure*}[!htbp]
    \centering
    \includegraphics[width=0.98\textwidth]{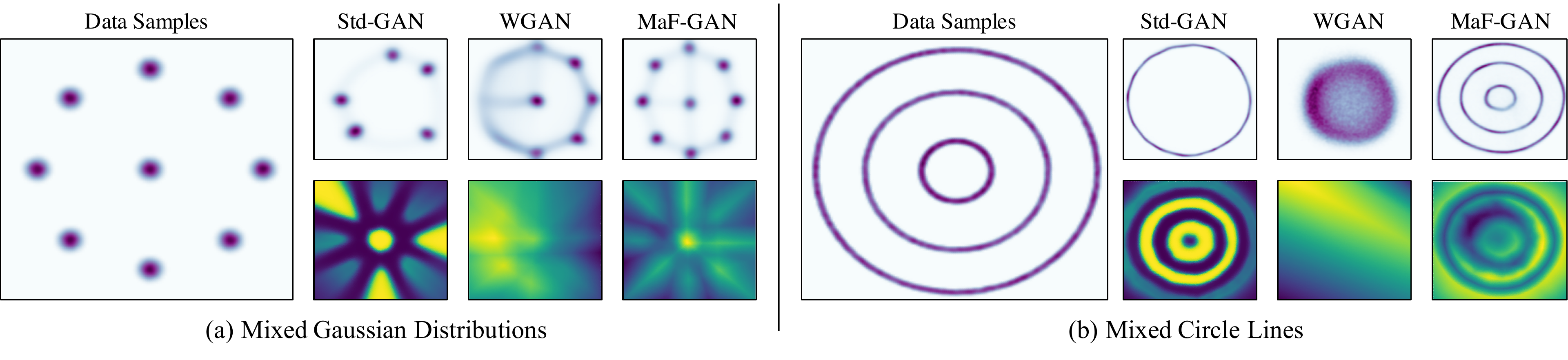}
    \caption{The experimental results carried on the synthetic data set, (a) 2D points from the mixture of 9 Gaussian distributions. From left to right: the samples generated by Std-GAN, WGAN and MaF-GAN. The first row refers to the generated results and the second row is the corresponding confidence map. (b)  three circle lines. }
    \label{fig:toys}
\end{figure*}

\textbf{Synthetic Dataset} consists of data from two different distributions, including mixed Gaussian distribution \cite{papamakarios2017masked} and mixed circle lines \cite{behrmann2019invertible}. As shown in Fig.\ref{fig:toys}, 50k points are sampled from the distribution and each point is represented as a vector containing abscissa and ordinate values.  $G(\cdot)$ consists of 4 fully-connected hidden layers and  $D(\cdot)$ is composed of three fully-connected layers. ReLU activation and batch normalization are used in $G(\cdot)$, while LinearMaxout without any batch normalization is adopted in $D(\cdot)$. The input code $z$ is a 32-dimensional vector sampled from a standard normal distribution. All models are trained by Adam \cite{kingma2014adam} for 500 epochs.



\begin{table*}[]
\setlength\tabcolsep{20pt}
\caption{The ablation study of the proposed method on CIFAR10 in terms of FID. Mean $\pm$ S.d. refers to the statistics of FID scores when the models are trained by 100, 400 and 800 epochs. Min is the optimal result obtained through training. }
\label{tab:ablation}
\centering
\resizebox{1.\textwidth}{!}{
\begin{tabular}{c|ccc|c|c}
\hline
                 & Epoch-100 & Epoch-400 & Epoch-800 & \textbf{Mean $\pm$ S.d.} & \textbf{Min}   \\ \hline
WGAN             & 133.54     &  106.71     &    80.17   &   106.81 $\pm$ 21.79 &   55.96   \\ \hline \hline
MaF-CGAN w/o T.C.      & 184.29              & 71.20       &    57.51        &     104.33 $\pm$   56.81    &    54.97   \\ \hline
\rowcolor[HTML]{EFEFEF}
MaF-CGAN    & \textbf{105.95}     & \textbf{58.60}           &    \textbf{50.27}        &  \textbf{71.61 $\pm$ 24.52}             &   \textbf{39.24}    \\ \hline\hline
MaF-DGAN w/o T.C.    & 149.90              & 80.99       &    61.44         &   97.44 $\pm$ 37.94            &   52.25    \\ \hline
\rowcolor[HTML]{EFEFEF}
MaF-DGAN  & \textbf{64.24}      & \textbf{41.47}       & \textbf{36.88}      &  \textbf{47.53 $\pm$ 11.96}             & \textbf{33.73} \\ \hline \hline
MaF-$\mathbb{E}$GAN w/o Entropy \& T.C.   & 104.03     &   60.86     &    50.90   &   71.93  $\pm$ 23.06 &   47.69   \\ \hline
MaF-$\mathbb{E}$GAN w/o T.C.           & 106.84     &   60.80     &    51.17   &   72.94  $\pm$ 24.29 &   46.66   \\ \hline
\rowcolor[HTML]{EFEFEF}
MaF-$\mathbb{E}$GAN           & \textbf{50.44}      & \textbf{35.83}  & \textbf{32.30}&  \textbf{39.52 $\pm$ 7.85}&  \textbf{30.85} \\
\end{tabular}
}
\end{table*}
\begin{figure*}[!htbp]
    \centering
    \includegraphics[width=1.\textwidth]{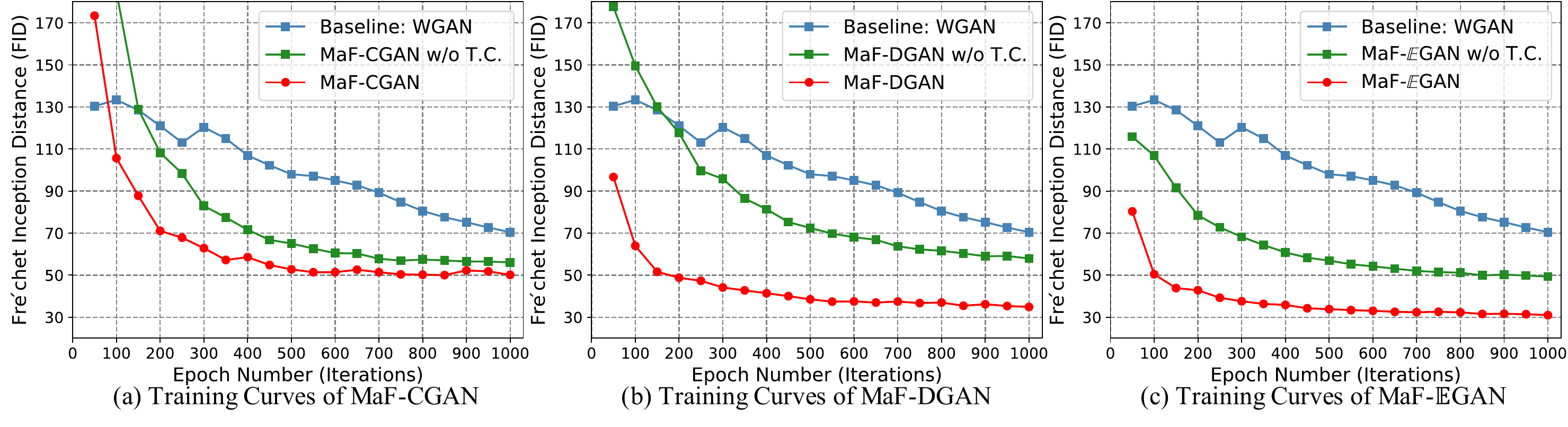}
    \caption{The experimental results carried on CIFAR 10. (a) shows the FIDs of Maf-CGAN for different training epochs. (b) the results of MaF-DGAN. (c) the results of MaF-$\mathbb{E}$GAN.  }
    \label{fig:training curves}
\end{figure*}

\textbf{CIFAR10} consists of 60,000 images with 10 classes (6000 images per class).
In this case, DCGAN \cite{radford2015unsupervised} is selected as the architecture to test the performance of different learning objectives. The model is trained by Adam with $\beta_1 $=0.0 and $\beta_2$=0.999. The learning rate is 0.0001 with a decay rate 0.9 for every 50 epochs. Batch size for training is 64. A 64-dimensional Gaussian distribution is adopted as the input for $G(\cdot)$, while the output of $f(D(\cdot))$ is set as a 16-dimensional embedding code. By following the most recent protocol \cite{xiangli2020real}, batch normalization \cite{ioffe2015batch} is used in $G(\cdot)$, while spectral normalization \cite{miyato2018spectral} is used in $D(\cdot)$. $D(\cdot)$ is updated for 3 times per $G(\cdot)$'s update.

To quantify the generation performance of the different methods, Fr\'{e}chet Inception Distance (FID) \cite{heusel2017gans} is adopted as the metric. FID computes the Wasserstein-2 distance between $p_g$ and $p_r$, which is a comprehensive and solid metric. FID with lower value refers to better generetive results. In all experiments, 50,000 images are randomly sampled to calculate FID. The implementation of FID in this work is based on the code available at \url{https://github.com/mseitzer/pytorch-fid}

\textbf{CelebA} consists of 202,599 face images labeled with 40 facial attributes. The images are cropped, aligned and resized to 256 $\times$ 256. The learning rate is 0.0001 with a decay rate 0.9 per 2 epochs.  A 128-dimensional Gaussian distribution is adopted as the input for $G(\cdot)$, and the output of $f(D(\cdot))$ is set as a 32-dimensional embedding code. $D(\cdot)$ and $G(\cdot)$ are updated step by step. Remaining settings including architecture, optimizer and evaluation metric, are identical with the setting for CIFAR10.

\textbf{Competing Methods.} Since only learning objective is redesigned in the proposed method, we mainly compare the proposed method to other popular objectives for GANs, including standard GAN(Std-GAN) \cite{goodfellow2014generative}, WGAN \cite{arjovsky2017wasserstein}, WGAN-GP \cite{gulrajani2017improved}, HingeGAN \cite{zhao2016energy}, LSGAN \cite{mao2017least} and Realness GAN \cite{xiangli2020real}. Note that Realness GAN is the most recent study mapping $g_p$ and $g_r$ into high-dimensional latent space. However this method does not pay attention to the preservation of topological structure. Hence, realness GAN is an ideal baseline for MaF-GANs to further prove the effectiveness of  Topological Consistency. Meanwhile, the proposed method is based on Wasserstein distance, therefore WGAN is another crucial baseline, which can help to clarify the contribution of better representation provided by  $D(\cdot)$ .

The public platform PyTorch \cite{paszke2017automatic} is used for implementation of the experiments on a work station with CPU of 2.8GHz, RAM of 512GB and GPUs of NVIDIA Tesla V100.

\subsection{Effectiveness Analysis of the Proposed Method}
\textbf{Ablation study.} To quantify the contribution of each component in MaF-GANs, we test the discriminative performance of the variants with or without the components. Table \ref{tab:ablation} shows the FID scores of different methods trained with different epochs on CIFAR10.
Since the proposed method is based on Wasserstein distance, the baseline of MaF-GANs is set to WGAN. As the results listed on Table \ref{tab:ablation}, the proposed methods,  including MaF-CGAN, MaF-DGAN and MaF-$\mathbb{E}$GAN, outperform the baselines significantly among all cases. When Topological Consistency is not adopted, the proposed  MaF-$\mathbb{E}$GAN can achieves 46.66 FID, which outperforms WGAN (55.96) by a large margin. Such improvement indicates that embedding data $x$ into a high-dimensional discriminative embedding space can boost the performance of $D(\cdot)$ effectively.

To further investigate the effectiveness of Topological Consistency, we conduct experiments to record the intermediate FID results of baseline and MaF-GAN with or without Topological Consistency on CIFAR10.
By observing the training curves shown in Fig. \ref{fig:training curves}, one can learn that great improvements ($\sim$30\% FID decrease) are achieved by adopting Topological Consistency.

As MaF-$\mathbb{E}$GAN achieves the best FID score shown in Table \ref{tab:ablation}, we adopt MaF-$\mathbb{E}$GAN as a representation of the zoo of MaF-GAN for the following experiments unless noted otherwise.

\textbf{Discussion on metrics for Topological Consistency}
In this section, we discuss the performance of different metrics applied for Topological Consistency.
Based on the theoretical analysis in Section. \ref{sec:ta}, several metrics can be used for Topological Consistency. To determine a more reasonable metric, three metrics, including L1 distance, Cosine Similarity and Mean Square Error, are considered in this paper. As the results listed on Table \ref{tab:metric}, when Mean Square Error is used , MaF-GAN achieves 30.85 FID, which is the best results among different metrics. Therefore, mean square error is selected as the metric for Topological Consistency and adopted for the other experiments.
\begin{figure*}[!htbp]
    \centering
    \includegraphics[width=1.\textwidth]{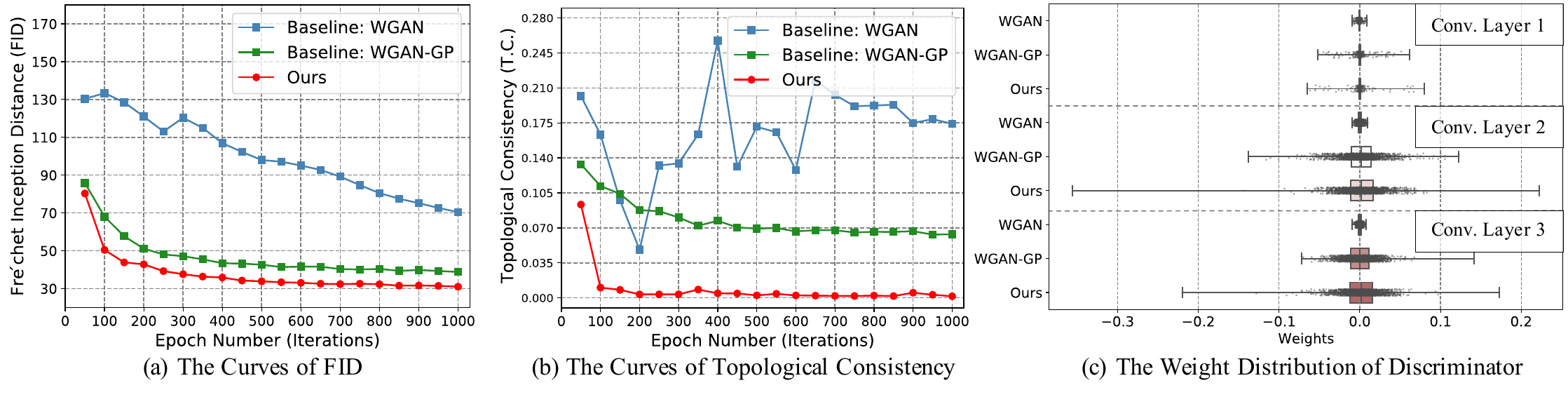}
    \caption{The experimental results carried on CIFAR 10. (a) the change of FIDs of WGAN, WGAN-GP and ours during training. (b) the values of Topological Consistency for different methods.  (c) the weight distribution of different layers in $D(\cdot)$ (Epoch=800). }
    \label{fig:topological}
\end{figure*}
\begin{table*}[]
\centering
\caption{The FID scores of the proposed MaF-GAN with different metrics on Topological Consistency on CIFAR10. }
\label{tab:metric}
\setlength\tabcolsep{20pt}
\resizebox{1.\textwidth}{!}{
\begin{tabular}{c|ccc|c|l}
\hline
Metric on T.C.    & Epoch-100      & Epoch-400      & Epoch-800      & \textbf{Mean $\pm$ S.d.}  & \textbf{Min}   \\ \hline
L1 Distance       & 65.74          & 42.48          & 35.43          & 47.88 $\pm$ 12.95         & 34.73          \\
Cosine Similarity & 50.61          & 37.17          & 33.28          & 40.35 $\pm$ 7.42          & 32.81          \\ \hline
\rowcolor[HTML]{EFEFEF}
Mean Square Error & \textbf{50.44} & \textbf{35.83} & \textbf{32.30} & \textbf{35.92 $\pm$ 7.85} & \textbf{30.85} \\ \hline
\end{tabular}
}
\end{table*}
\begin{table}[]
\centering
\caption{The FID scores of the proposed MaF-GAN for $\mathbf{V}_x$ with different lengths on CIFAR10 }
\label{tab:outcomes}
\resizebox{.48\textwidth}{!}{
\begin{tabular}{c|cccc}
\hline
The length of $\mathbf{V}_x$ & Epoch-1           & Epoch-50       & Epoch-100       & Epoch-150      \\ \hline
8        & \textbf{359.71}     &                      \underline{78.90}  &\textbf{49.40}& \underline{44.05}         \\
16       & 502.34     &               80.00         & 50.44  & \textbf{43.98}          \\
32       & 414.37     &\textbf{75.84}&  \underline{50.30}          & 44.06          \\
64       & \underline{384.48}     &  84.09         &  52.56          & 45.77          \\ \hline
\rowcolor[HTML]{EFEFEF}
S.d.     &  53.89              &  2.95          &  1.16           &   0.75             \\ \hline
\end{tabular}
}
\end{table}

\textbf{The robustness against hyper-parameters}
In this section, we discuss the robustness of the proposed method against various hyper-parameters on CIFAR10. We first evaluate the performance of MaF-GANs on FID with different $\mathbf{V}_x \in R^n$, including $n = 8, 16, 32 \; and \; 64$. And then we conduct experiments on constant $K$ of Lipschitz constraint to further prove Theorem 1 experimentally, including $K = 1, 5\; \text{and} \; 10$.

We now discuss the experimental results of different $n$. As shown in Table \ref{tab:outcomes}, the experimental results show that the FID scores gradually converge to similar values.
Most importantly, the standard deviation of the FID scores is 0.75 at Epoch-150, which indicates that different length of $\mathbf{V}_x$ would not make huge difference to the performance of generator. Though larger $n$ introduces more redundancy, the proposed Maf-GAN remains a great level of robustness.


We now evaluate the performance of MaF-GANs under different constant $K$ of Lipschitz constraint.
Based on the \textbf{Proposition 1.}, $K$ is an absolute term for the objective of $D(\cdot)$, hence we can analyze the robustness of the proposed method against $K$ by changing the weights of $\mathbb{L}$ and $\mathbb{D}_{TC}$.
As proven in this paper, the proposed Topological Consistency is an universal solution for all real constant $K$ of Lipschitz constraint. To prove such superiority of the proposed method, we train several MaF-GANs with different $K$s and observe the generated results.
As listed in Table \ref{tab:KLC},  with the training of MaF-GAN, the FID scores eventually become very similar. In particular, for the Epoch-50 case, the variance of FID is 12.58, but for the Epoch-350, the variance decreases to 0.71, which further justifies the theory.
\begin{figure*}[!htbp]
    \centering
    \includegraphics[width=0.98\textwidth]{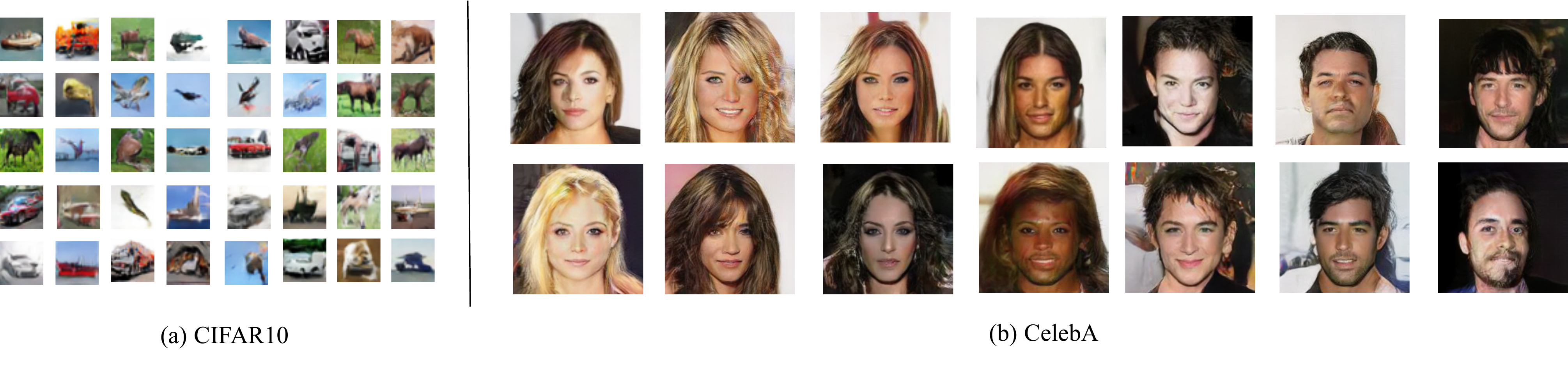}
    \caption{The experimental results carried on the real data set with DCGAN architecture. (a) refers to the case of CIFAR10 and (b) is the case of CelebA}
    \label{fig:realdata}
\end{figure*}

\subsection{Effect of Lipschitz continuity on the GANs based on Wasserstein Distance}

As an ideal metric for GANs, Wasserstein distance is widely used in some famous learning objectives, such as WGAN, WGAN-GP and MaF-GAN. However, the similar metric leads to the quite different generated results. In order to investigate the radical rational for such phenomenon, Topological Consistency is drawn as an analysis tool to observe the continuity of $D(\cdot)$ during training. As shown in Fig.\ref{fig:topological}, we visualize the FID, Topological Consistency and the weight distribution separately. By observing the FID scores, MaF-GAN is the winner in the speed of convergence and the realness of generated images. Compared with Topological Consistency, WGAN-GP can achieve similar speed but lower realness. For WGAN, a wide margin can be seen in
\begin{table}[]
\centering
\caption{The FID scores of the proposed MaF-GAN with different $K$ for Lipschitz continuity on CIFAR10}
\label{tab:KLC}
\large
\resizebox{.48\textwidth}{!}{
\begin{tabular}{c|cccc}
\hline
\begin{tabular}[c]{@{}c@{}}$K$ for \\ Lipschitz continuity\end{tabular} & Epoch-50       & Epoch-150      & Epoch-250      & Epoch-350      \\ \hline
1                            & 80.00          & \textbf{43.98} & \textbf{39.25} & \textbf{36.33} \\
5                            & \underline{77.64}    & 46.58          & \underline{41.28}    & \underline{37.58}    \\
10                           & \textbf{52.21} & \underline{45.21}    & 41.31          & 38.00          \\ \hline
\rowcolor[HTML]{EFEFEF}
S.d.                         & 12.58          & 1.06           & 0.96           & 0.71           \\ \hline
\end{tabular}
}
\end{table}
Fig. \ref{fig:topological}(a), which indicates that the realness of WGAN is quite lower than other methods. Such experimental results can be clearly explained by the variance of Topological Consistency. As shown in Fig. \ref{fig:topological}(b), the output of the second last fully-connected layer in $D(\cdot)$ is adopted to calculate Topological Consistency. WGAN performs a significant fluctuation in Topological Consistency, but the trend of steadiness can be observed in WGAN-GP and MaF-GAN.  Weight clipping seems not to be a reasonable way for Lipschitz continuity, as it leads to a large variances in the terms of continuity. Such property of weight clipping is the radical result for the hard training of WGAN. On the other side, Topological Consistency of WGAN-GP is a solid evidence to prove the effectiveness of the proposed method. Without taking Topological Consistency as learning objective, WGAN-GP also reaches a very competitive results in Fig.\ref{fig:topological}(b). This indicates that Topological Consistency can estimate the smooth of $D(\cdot)$ precisely and the degree of continuity is the key factor for the generated results. Note that, existing methods pay more attention to limiting the parameter space, however, the results shown in Fig. \ref{fig:topological}(c) illustrate that MaF-GAN provides the largest parameter space but the least sharp variance in $D(\cdot)$, which effectively proves the correctness of the learning objective in MaF-GAN.

When comes to the cost of training time, Topological Consistency also outperforms gradient penalty. As the time cost listed in Table \ref{tab:time}, By adopting Topological Consistency as the Lipschitz constraint, only 31.53s is required to train GAN, which exceeds WGAN-GP by 35.28\% absolutely. This empirically indicates that the proposed method not only achieves better results but also works in a more efficient way.

\begin{table}[]
\centering
\caption{The time cost (s) to train the proposed method and WGAN-GP on CIFAR10}
\label{tab:time}
\resizebox{.48\textwidth}{!}{
\begin{tabular}{c|cc|c}
\hline
         & \begin{tabular}[c]{@{}c@{}}Training $D(\cdot)$ \\ with $\mathcal{L}$\end{tabular} & \begin{tabular}[c]{@{}c@{}}Training $D(\cdot)$ \\ with Lipschitz Constraint\end{tabular} & \textbf{\begin{tabular}[c]{@{}c@{}}Time Cost \\ per Epoch\end{tabular}} \\ \hline
WGAN-GP              & \textbf{0.0147}                                                                   & 0.0219                                                                                   & 36.15                                                                   \\ \hline
\rowcolor[HTML]{EFEFEF}
MaF-GAN    & 0.0165                                                                            & \textbf{0.0169}                                                                          & \textbf{31.53}                                                          \\ \hline
\end{tabular}
}
\end{table}
\begin{table}[]
\centering
\caption{ FIDs on CelebA and CIFAR10 for DCGAN with various learning objectives}
\label{tab:compared}
\resizebox{.5\textwidth}{!}{
\setlength\tabcolsep{10pt}
\begin{tabular}{c|cc}
\hline
Learning Objective             & CIFAR-10                       & CelebA                 \\ \hline
WGAN                           & 55.96                          &   -                    \\
HingeGAN                       & 42.40                          & 25.57                  \\
LSGAN                          & 42.01                           & 30.76                  \\
DCGAN                          & 38,56                          & 27.02                  \\ \hline
WGAN-GP                        & 41.86                          & 70.28                  \\
WGAN-GP reported in this paper & 38.63                          & 70.16                  \\ \hline
Realness GAN-Obj.1             & 36.73                          & -                      \\
Realness GAN-Obj.2             & 34.59                          & \underline{23.51}                  \\
Realness GAN-Obj.3             & 36.21                          & -                      \\ \hline\hline
\rowcolor[HTML]{EFEFEF}
Ours: CwGAN-T.C.        & 39.24                          &          -              \\
\rowcolor[HTML]{EFEFEF}
Ours: DwGAN-T.C.        & \underline{33.73}              &          -              \\
\rowcolor[HTML]{EFEFEF}
Ours: $\mathbb{E}$wGAN-T.C.                & \textbf{30.85}                 &     \textbf{12.43}                   \\ \hline
\end{tabular}
}
\end{table}

\subsection{Comparing with Existing Methods}
To further verify the effectiveness of the proposed method, we compare it with widely-used learning objectives, i.e. Std-GAN, WGAN, HingeGAN, LSGAN, DCGAN, WGAN-GP and Realness GAN. In particular, a synthetic dataset and two real dataset, including CIFAR10 and CelebA, are considered.

As shown in Fig.\ref{fig:toys}, the proposed method shows strong superiority in estimating the complex two-dimensional distribution in each case. To show the advance of MaF-GAN over Std-GAN and WGAN, we visualize the confidence map of discriminator. In the confidence map, higher value indicates that discriminator more confidently judges the selected point as real sample. Through the confidence map, it can be seen that Std-GAN maps the real data sample into an imbalance mode, where some real lines or points are associated with very low confidence and others are lied on pretty high values. The real points or lines with strong confidence are hard for generator to search and learn, resulting in the mode collapse. When comes to WGAN, over-continuous confidence map is provided by discriminator, which will contribute some confusing directions for generator, leading to gradient exploding. The confidence map of discriminator is crucial for GANs. To draw the precise and gradual variance in the map, the proposed method embeds the data into a high-dimensional latent space with Topological Consistency. With more capacity to evaluate the realness, the proposed method can derive a more reasonable confidence map, which finally boosts the generation performance.

To further show the superiority of the proposed method, we compare the performance of the proposed method with the state-of-the-art methods on CIFAR10 and CelebA. As the results shown in Table \ref{tab:compared}, the proposed method outperforms other methods in both datasets by a wide margin. Compared with the most recent method, Realness GAN, a 10.81\% improvement in FID is achieved by the proposed method on CIFAR10. Similarly, in the case of CelebA, the FIDs of Realness GAN and the proposed method are 23.51 and 12.43, respectively, which convincingly shows the advantage of MaF-GAN. Note that the results reported on Table \ref{tab:compared}  are all based on the same architecture, i.e. DCGAN.
\section{Conclusion}
In this paper, we propose Manifold-preserved GANs, denoted as MaF-GANs, to mitigate mode collapse and gradient exploding for the training of GANs. Unlike existing learning objectives, MaF-GANs first map the real/generated data into a high-dimensional embedding space, and judge the realness of the samples from a manifold-based view. To leverage more progressive gradient, Topological Consistency, which lands on the manifold, is then proposed to preserve the topological structure via embedding. Finally, the diversity and realness of generated data is facilitated by the stronger representation and flatter embedding space. We theoretically prove the advantage of the proposed method by analyzing the equivalence between Lipschitz continuity and Topological Consistency. Meanwhile, extensive experiments on various datasets with different sizes, i.e., a synthetic dataset (2*1), CIFAR10 (32*32) and CelebA (256*256) are carried out to further validate the effectiveness of the proposed method.
In the future, we will further investigate the application of the proposed method in other open-set problems, such as open-set classification and self-supervised learning.
\section*{Acknowledgement}
The work is partially supported by the National Natural Science Foundation of China under grants no. 62076163 and 91959108, the Shenzhen Fundamental Research fund JCYJ20190808163401646, JCYJ20180305125822769, Tencent “Rhinoceros Birds”-Scientific Research Foundation for Young Teachers of Shenzhen University.
\bibliographystyle{IEEEtran}
\bibliography{myreference}

\begin{thebibliography}{10}
\providecommand{\url}[1]{#1}
\csname url@samestyle\endcsname
\providecommand{\newblock}{\relax}
\providecommand{\bibinfo}[2]{#2}
\providecommand{\BIBentrySTDinterwordspacing}{\spaceskip=0pt\relax}
\providecommand{\BIBentryALTinterwordstretchfactor}{4}
\providecommand{\BIBentryALTinterwordspacing}{\spaceskip=\fontdimen2\font plus
\BIBentryALTinterwordstretchfactor\fontdimen3\font minus
  \fontdimen4\font\relax}
\providecommand{\BIBforeignlanguage}[2]{{%
\expandafter\ifx\csname l@#1\endcsname\relax
\typeout{** WARNING: IEEEtran.bst: No hyphenation pattern has been}%
\typeout{** loaded for the language `#1'. Using the pattern for}%
\typeout{** the default language instead.}%
\else
\language=\csname l@#1\endcsname
\fi
#2}}
\providecommand{\BIBdecl}{\relax}
\BIBdecl

\bibitem{creswell2018generative}
A.~Creswell, T.~White, V.~Dumoulin, K.~Arulkumaran, B.~Sengupta, and A.~A.
  Bharath, ``Generative adversarial networks: An overview,'' \emph{IEEE Signal
  Processing Magazine}, vol.~35, no.~1, pp. 53--65, 2018.

\bibitem{8411144}
H.~Zhang, T.~Xu, H.~Li, S.~Zhang, X.~Wang, X.~Huang, and D.~N. Metaxas,
  ``Stackgan++: Realistic image synthesis with stacked generative adversarial
  networks,'' \emph{IEEE Transactions on Pattern Analysis and Machine
  Intelligence}, vol.~41, no.~8, pp. 1947--1962, 2019.

\bibitem{8968618}
J.~Pan, J.~Dong, Y.~Liu, J.~Zhang, J.~Ren, J.~Tang, Y.-W. Tai, and M.-H. Yang,
  ``Physics-based generative adversarial models for image restoration and
  beyond,'' \emph{IEEE Transactions on Pattern Analysis and Machine
  Intelligence}, vol.~43, no.~7, pp. 2449--2462, 2021.

\bibitem{9149832}
J.~Cao, Y.~Guo, Q.~Wu, C.~Shen, J.~Huang, and M.~Tan, ``Improving generative
  adversarial networks with local coordinate coding,'' \emph{IEEE Transactions
  on Pattern Analysis and Machine Intelligence}, pp. 1--1, 2020.

\bibitem{9496081}
Y.~Tian, L.~Shen, L.~Shen, G.~Su, Z.~Li, and W.~Liu, ``Alphagan: Fully
  differentiable architecture search for generative adversarial networks,''
  \emph{IEEE Transactions on Pattern Analysis and Machine Intelligence}, pp.
  1--1, 2021.

\bibitem{liu2021group}
H.~Liu, H.~Wu, W.~Xie, F.~Liu, and L.~Shen, ``Group-wise inhibition based
  feature regularization for robust classification,'' \emph{arXiv preprint
  arXiv:2103.02152}, 2021.

\bibitem{arjovsky2017wasserstein}
M.~Arjovsky, S.~Chintala, and L.~Bottou, ``Wasserstein generative adversarial
  networks,'' in \emph{International conference on machine learning}.\hskip 1em
  plus 0.5em minus 0.4em\relax PMLR, 2017, pp. 214--223.

\bibitem{gulrajani2017improved}
I.~Gulrajani, F.~Ahmed, M.~Arjovsky, V.~Dumoulin, and A.~Courville, ``Improved
  training of wasserstein gans,'' \emph{arXiv preprint arXiv:1704.00028}, 2017.

\bibitem{wu2018wasserstein}
J.~Wu, Z.~Huang, J.~Thoma, D.~Acharya, and L.~Van~Gool, ``Wasserstein
  divergence for gans,'' in \emph{Proceedings of the European Conference on
  Computer Vision (ECCV)}, 2018, pp. 653--668.

\bibitem{stanczuk2021wasserstein}
J.~Stanczuk, C.~Etmann, L.~M. Kreusser, and C.-B. Sch{\"o}nlieb, ``Wasserstein
  gans work because they fail (to approximate the wasserstein distance),''
  \emph{arXiv preprint arXiv:2103.01678}, 2021.

\bibitem{karras2017progressive}
T.~Karras, T.~Aila, S.~Laine, and J.~Lehtinen, ``Progressive growing of gans
  for improved quality, stability, and variation,'' in \emph{Proceedings of the
  International Conference on Learning Representations (ICLR)}, 2017.

\bibitem{karras2019style}
T.~Karras, S.~Laine, and T.~Aila, ``A style-based generator architecture for
  generative adversarial networks,'' in \emph{Proceedings of the IEEE/CVF
  Conference on Computer Vision and Pattern Recognition}, 2019, pp. 4401--4410.

\bibitem{choi2018stargan}
Y.~Choi, M.~Choi, M.~Kim, J.-W. Ha, S.~Kim, and J.~Choo, ``Stargan: Unified
  generative adversarial networks for multi-domain image-to-image
  translation,'' in \emph{Proceedings of the IEEE conference on computer vision
  and pattern recognition}, 2018, pp. 8789--8797.

\bibitem{goodfellow2014generative}
I.~Goodfellow, J.~Pouget-Abadie, M.~Mirza, B.~Xu, D.~Warde-Farley, S.~Ozair,
  A.~Courville, and Y.~Bengio, ``Generative adversarial nets,'' \emph{Advances
  in neural information processing systems}, vol.~27, 2014.

\bibitem{mao2017least}
X.~Mao, Q.~Li, H.~Xie, R.~Y. Lau, Z.~Wang, and S.~Paul~Smolley, ``Least squares
  generative adversarial networks,'' in \emph{Proceedings of the IEEE
  international conference on computer vision}, 2017, pp. 2794--2802.

\bibitem{zhao2016energy}
J.~Zhao, M.~Mathieu, and Y.~LeCun, ``Energy-based generative adversarial
  network,'' in \emph{Proceedings of the International Conference on Learning
  Representations (ICLR)}, 2017.

\bibitem{villani2009optimal}
C.~Villani, \emph{Optimal transport: old and new}.\hskip 1em plus 0.5em minus
  0.4em\relax Springer, 2009, vol. 338.

\bibitem{krizhevsky2009learning}
A.~Krizhevsky, G.~Hinton \emph{et~al.}, ``Learning multiple layers of features
  from tiny images,'' 2009.

\bibitem{liu2015deep}
Z.~Liu, P.~Luo, X.~Wang, and X.~Tang, ``Deep learning face attributes in the
  wild,'' in \emph{Proceedings of the IEEE international conference on computer
  vision}, 2015, pp. 3730--3738.

\bibitem{papamakarios2017masked}
G.~Papamakarios, T.~Pavlakou, and I.~Murray, ``Masked autoregressive flow for
  density estimation,'' \emph{arXiv preprint arXiv:1705.07057}, 2017.

\bibitem{behrmann2019invertible}
J.~Behrmann, W.~Grathwohl, R.~T. Chen, D.~Duvenaud, and J.-H. Jacobsen,
  ``Invertible residual networks,'' in \emph{International Conference on
  Machine Learning}.\hskip 1em plus 0.5em minus 0.4em\relax PMLR, 2019, pp.
  573--582.

\bibitem{kingma2014adam}
D.~P. Kingma and J.~Ba, ``Adam: A method for stochastic optimization,''
  \emph{arXiv preprint arXiv:1412.6980}, 2014.

\bibitem{radford2015unsupervised}
A.~Radford, L.~Metz, and S.~Chintala, ``Unsupervised representation learning
  with deep convolutional generative adversarial networks,'' \emph{arXiv
  preprint arXiv:1511.06434}, 2015.

\bibitem{xiangli2020real}
Y.~Xiangli, Y.~Deng, B.~Dai, C.~C. Loy, and D.~Lin, ``Real or not real, that is
  the question,'' \emph{International Conference on Learning Representations
  (ICLR)}, 2020.

\bibitem{ioffe2015batch}
S.~Ioffe and C.~Szegedy, ``Batch normalization: Accelerating deep network
  training by reducing internal covariate shift,'' in \emph{International
  conference on machine learning}.\hskip 1em plus 0.5em minus 0.4em\relax PMLR,
  2015, pp. 448--456.

\bibitem{miyato2018spectral}
T.~Miyato, T.~Kataoka, M.~Koyama, and Y.~Yoshida, ``Spectral normalization for
  generative adversarial networks,'' \emph{arXiv preprint arXiv:1802.05957},
  2018.

\bibitem{heusel2017gans}
M.~Heusel, H.~Ramsauer, T.~Unterthiner, B.~Nessler, and S.~Hochreiter, ``Gans
  trained by a two time-scale update rule converge to a local nash
  equilibrium,'' \emph{Advances in neural information processing systems},
  vol.~30, 2017.

\bibitem{paszke2017automatic}
A.~Paszke, S.~Gross, S.~Chintala, G.~Chanan, E.~Yang, Z.~DeVito, Z.~Lin,
  A.~Desmaison, L.~Antiga, and A.~Lerer, ``Automatic differentiation in
  pytorch,'' 2017.

\end{thebibliography}
\end{document}